\documentclass[10pt,twocolumn,letterpaper]{article}

\makeatother

\usepackage{cvpr}              % To produce the CAMERA-READY version

\usepackage[accsupp]{axessibility}

\usepackage{graphicx}
\usepackage{amsmath}
\usepackage{amssymb}
\usepackage{booktabs}

\usepackage{algorithmicx}
\usepackage[ruled]{algorithm}
\usepackage[noend]{algpseudocode}

\usepackage{multirow, makecell}

\newcommand\cred[1]{\textbf{\textcolor{red}{#1}}} 
\newcommand\cblue[1]{\textbf{\textcolor{blue}{#1}}}

\usepackage{wrapfig}
\usepackage{bm}
\usepackage[utf8]{inputenc} % allow utf-8 input
\usepackage[T1]{fontenc}    % use 8-bit T1 fonts
\usepackage{url}            % simple URL typesetting
\usepackage{amsfonts}       % blackboard math symbols
\usepackage{nicefrac}       % compact symbols for 1/2, etc.
\usepackage{microtype}      % microtypography
\usepackage{xcolor}         % colors

\usepackage[pagebackref,breaklinks,colorlinks]{hyperref}

\usepackage[capitalize]{cleveref}
\crefname{section}{Sec.}{Secs.}
\Crefname{section}{Section}{Sections}
\Crefname{table}{Table}{Tables}
\crefname{table}{Tab.}{Tabs.}

\def\cvprPaperID{756} % *** Enter the CVPR Paper ID here
\def\confName{CVPR}
\def\confYear{2023}

\begin{document}

\title{Progressive Random Convolutions for Single Domain Generalization}

\author{
Seokeon Choi~~~ 
Debasmit Das~~~ 
Sungha Choi~~~ 
Seunghan Yang~~~ 
Hyunsin Park~~~ 
Sungrack Yun~~~
\smallskip
\\
Qualcomm AI research$^{\dagger}$
\\
\smallskip
{\tt\small \{seokchoi,debadas,sunghac,seunghan,hyunsinp,sungrack\}@qti.qualcomm.com}
}
\maketitle

\begin{abstract}

Single domain generalization aims to train a generalizable model with only one source domain to perform well on arbitrary unseen target domains. Image augmentation based on Random Convolutions (RandConv), consisting of one convolution layer randomly initialized for each mini-batch, enables the model to learn generalizable visual representations by distorting local textures despite its simple and lightweight structure. However, RandConv has structural limitations in that the generated image easily loses semantics as the kernel size increases, and lacks the inherent diversity of a single convolution operation. To solve the problem, we propose a Progressive Random Convolution (Pro-RandConv) method that recursively stacks random convolution layers with a small kernel size instead of increasing the kernel size. This progressive approach can not only mitigate semantic distortions by reducing the influence of pixels away from the center in the theoretical receptive field, but also create more effective virtual domains by gradually increasing the style diversity. In addition, we develop a basic random convolution layer into a random convolution block including deformable offsets and affine transformation to support texture and contrast diversification, both of which are also randomly initialized. Without complex generators or adversarial learning, we demonstrate that our simple yet effective augmentation strategy outperforms state-of-the-art methods on single domain generalization benchmarks.

\end{abstract}

\begin{figure}[t!]
\centering
\includegraphics[width=\linewidth]{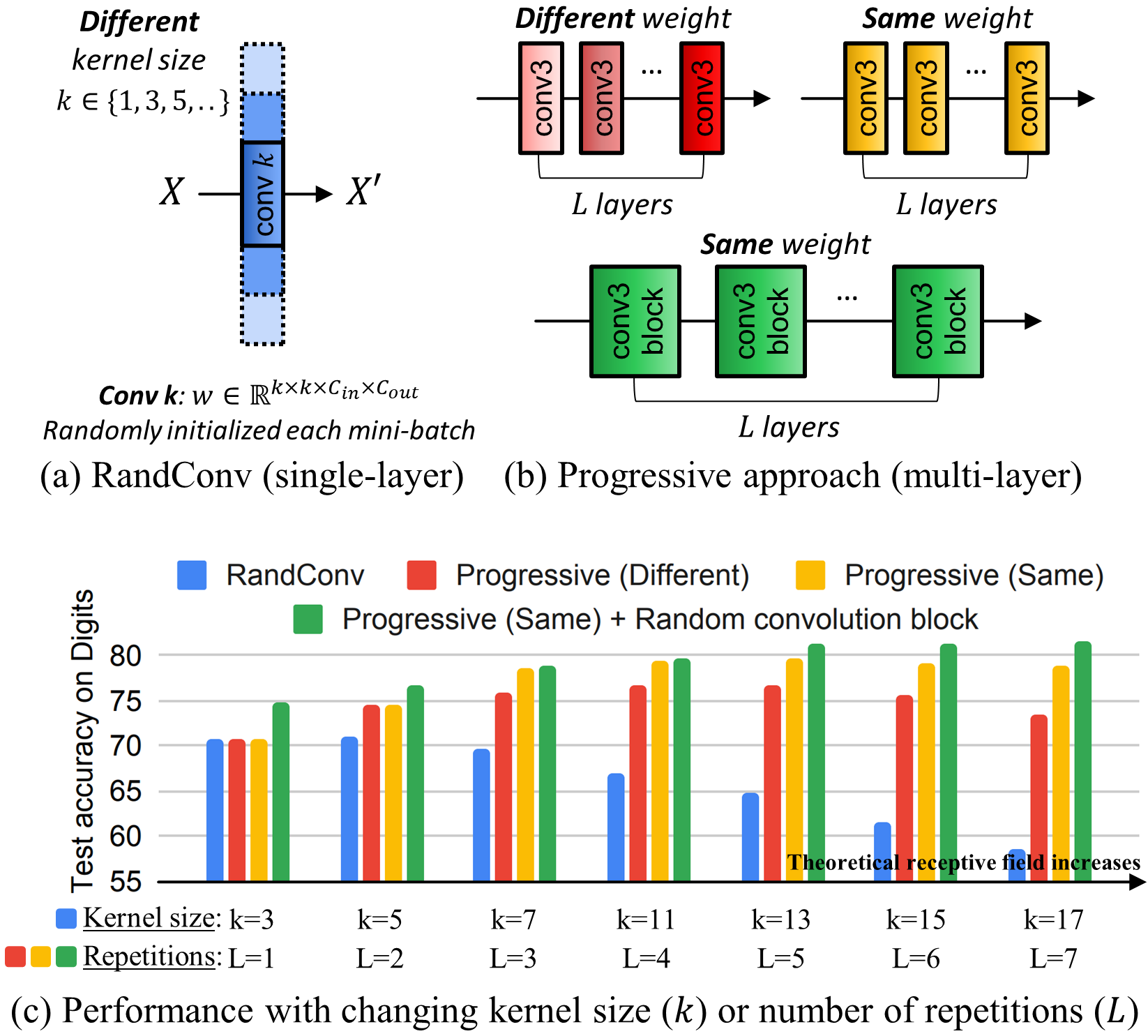}
\vspace{-0.3cm}
\caption{Comparison of conventional random convolutions (single-layer) and our progressive random convolutions (multi-layer). Our final model includes multiple random convolution blocks consisting of deformable offsets and affine transformation.}
\label{fig:concept}
\vspace{-0.3cm}
\end{figure}

\section{Introduction}

{\let\thefootnote\relax\footnotetext{{
\hspace{-6.5mm} $\dagger$ Qualcomm AI Research is an initiative of Qualcomm Technologies, Inc.}}}

\begin{figure*}[t!]
\centering
\vspace{-0.7cm}
\includegraphics[width=\linewidth]{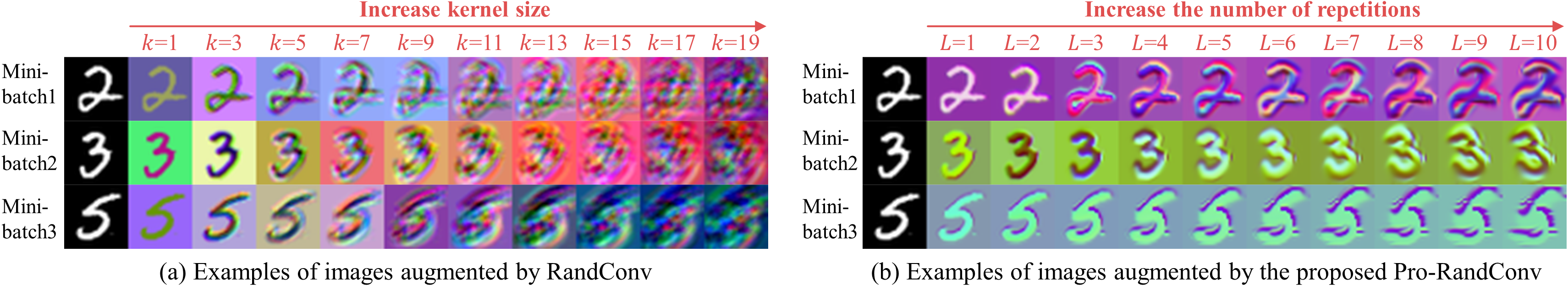}
\vspace{-0.45cm}
\caption{Examples of images augmented by RandConv and our Pro-RandConv composed of multiple convolution blocks.}
\label{fig:images}
\vspace{-0.35cm}
\end{figure*}

In recent years, deep neural networks have achieved remarkable performance in a wide range of applications~\cite{krizhevsky2012imagenet, lecun2015deep}. However, this success is built on the assumption that the test data (\ie target) should share the same distribution as the training data (\ie source), and they often fail to generalize to out-of-distribution data~\cite{taori2020measuring, djolonga2021robustness, hendrycks2021many}. In practice, this domain discrepancy problem between source and target domains is commonly encountered in real-world scenarios. Especially, a catastrophic safety issue may occur in medical imaging~\cite{yao2019strong, li2020domain} and autonomous driving~\cite{yang2018real, zhang2017curriculum} applications. To tackle this problem, one line of work focuses on domain adaptation (DA) to transfer knowledge from a source domain to a specific target domain~\cite{DANN, hoffman2018cycada, kim2019diversify, choi2022improving}. This approach usually takes into account the availability of labeled or unlabeled target domain data. Another line of work deals with a more realistic setting known as domain generalization (DG), which aims to learn a domain-agnostic feature representation with only data from source domains without access to target domain data. Thanks to its practicality, the task of domain generalization has been extensively studied. 

In general, the paradigm of domain generalization depends on the availability of using multi-source domains~\cite{wang2021learning}. Previously, many studies~\cite{muandet2013domain, ghifary2015domain, li2017deeper, grubinger2017multi, choi2021meta} have focused on using multi-source domains, and the distribution shift can be alleviated by simply aggregating data from multiple training domains~\cite{li2017deeper, li2019episodic}. However, this approach faces practical limitations due to data collection budgets~\cite{qiao2020learning}. As an alternative, the single domain generalization problem has recently received attention~\cite{li2021progressive, wang2021learning}, which learns robust representation using only a single source domain. A common solution to this challenging problem is to generate diverse samples in order to expand the coverage of the source domain through an adversarial data augmentation scheme~\cite{zhaoNIPS20maximum, NIPS2018_7779}. However, most of these methods share a complicated training pipeline with multiple objective functions. 

In contrast, Xu~\etal suggested Random Convolution (RandConv)~\cite{xu2020robust} that consists of a single convolution layer whose weights are randomly initialized for each mini-batch, as described in Fig.~\ref{fig:concept}(a). When RandConv is applied to an input image, it tends to modify the texture of the input image depending on the kernel size of the convolution layer. This is a simple and lightweight image augmentation technique compared to complex generators or adversarial data augmentation. Despite these advantages, this method has structural limitations. Firstly, the image augmented by RandConv easily loses its semantics while increasing the kernel size, which is shown in Fig.~\ref{fig:images}(a). As a result, the ability to generalize in the test domain is greatly reduced as shown in Fig.~\ref{fig:concept}(c). Secondly, RandConv lacks the inherent diversity of a single convolution operation.

To solve these limitations, we propose a progressive approach based on random convolutions, named Progressive Random Convolutions (Pro-RandConv). Figure~\ref{fig:concept}(b) describes the progressive approach consisting of multiple convolution layers with a small kernel size. Our progressive approach has two main properties. The first is that the multi-layer structure can alleviate the semantic distortion issues by reducing the impact on pixels away from the center in the theoretical receptive field, as revealed in~\cite{luo2016understanding}. Therefore, the progressive approach does not degrade the performance much even if the receptive field increases, as shown in Fig.~\ref{fig:concept}(c). The second property is that stacking random convolution layers of the same weights can generate more effective virtual domains rather than using different weights. This is an interesting observation that can be interpreted as gradually increasing the distortion magnitude of a single transformation to the central pixels. This approach enables more fine-grained control in image transformation than a single layer with a large kernel, which has the effect of incrementally improving style diversity.

In addition, we propose a random convolution block including deformable offsets and affine transformation to support texture and contrast diversification. It is noteworthy that all weights are also sampled from a Gaussian distribution, so our convolution block is an entirely stochastic process. Finally, we can maximize the diversity of styles while maintaining the semantics of newly generated images through the progressive method of this random convolution block, as described in Fig.~\ref{fig:images}(b). We argue that the proposed Pro-RandConv could be a strong baseline because it surpasses recent single DG methods only by image augmentation without an additional loss function or complex training pipelines. To summarize, our main contributions are as follows:

\vspace{-0.1cm}
{
\begin{itemize}
    \item We propose a progressive approach of recursively stacking small-scale random convolutions to improve the style diversity while preserving object semantics.
\vspace{-0.1cm}
    \item We develop a random convolution layer with deformable offsets and affine transformation to promote texture and contrast diversity for augmented images.
\vspace{-0.1cm}
    \item We perform comprehensive evaluation and analyses of our method on single and multi DG benchmarks on which we produce significant improvement in recognition performance compared to other methods.
\end{itemize}}

\section{Related work}

\subsection{Domain generalization}

Domain generalization aims to learn a model from source domains and generalize it to unseen target domains without access to any of the target domain data. Domain generalization is classified into single-source or multi-source depending on the number of source domains. Multi domain generalization methods tackle domain shift mostly by alignment~\cite{muandet2013domain,motiian2017few,li2018domain,dou2019domain} or ensembling~\cite{mancini2018best}. We note that these alignment and ensembling methods assume the existence of multiple source domains and cannot be trivially extended to the single domain generalization setup.

Single domain generalization methods rely on generating more diverse samples to expand the coverage of the source domain~\cite{NIPS2018_7779, qiao2020learning, tolstikhin2018wasserstein, zhaoNIPS20maximum,li2021progressive, wang2021learning}. These fictitious domains are generally created through an adversarial data augmentation scheme~\cite{NIPS2018_7779,qiao2020learning}. More recently, PDEN~\cite{li2021progressive} uses a progressive approach to generate novel domains by optimizing a contrastive loss function based on adversaries. In L2D~\cite{wang2021learning}, the authors propose to alternatively learn the stylization module and the task network by optimizing mutual information. Recently, Liu et al.~\cite{liu2022geometric} proposed to reduce the domain gap through geometric and textural augmentation by sampling from the distribution of augmentations for an object class. In MetaCNN~\cite{wan2022meta}, the authors propose to decompose convolutional features into meta features which are further processed to remove irrelevant components for better generalization. However, most of these methods have complex and dedicated training pipelines with multiple objective functions compared to our method which can generate new domains through simple augmentation.

\subsection{Data augmentation}

Data augmentation commonly uses manual methods for vision tasks including photometric transformations (\eg color jitter and grayscale) and geometric transformations (\eg translation, rotation, and shearing). To address challenges of manual data augmentation design, automatic augmentation methods~\cite{zhang2020generalizing,volpi2019addressing,Cubuk_2019_CVPR,cubuk2020randaugment,yang2022domain,yang2021distribution} have been proposed. AutoAugment~\cite{Cubuk_2019_CVPR} uses a reinforcement learning agent to optimize an augmentation policy while RandAugment~\cite{cubuk2020randaugment} decreases the search space of AutoAugment by randomly choosing a subset of augmentation type. However, these methods are not specialized enough to deal with large domain shifts present in the single domain generalization setting.

\subsection{Domain randomization}

Domain randomization, normally used for robotics and autonomous vehicles, varies the scene parameters in a simulation environment to expand the training data and produce better generalization. Tobin \etal~\cite{tobin2017domain} pioneered domain randomization by varying different configurations of the simulated environment for robot manipulation. Yue \etal~\cite{yue2019domain} extended domain randomization to semantic segmentation and proposed a pyramid consistency loss. Style randomization can also simulate novel source domains through random convolutions~\cite{xu2020robust} or normalization layer modulation~\cite{jackson2019style}. In our work, we also create new styles based on random convolutions~\cite{xu2020robust}, but we point out that there are inherent problems with the previous work. To solve these, we propose a progressive approach with a random convolution block, which can improve style diversity while maintaining semantics.

\section{Background}

\subsection{Problem formulation}

We first define the problem setting and notations. We assume that the source data is observed from a single domain $\mathcal{S}=\{\mathbf{x}_n,y_n\}_{n=1}^{N_S}$, where $\mathbf{x}_n$ and $y_n$ are the $n$-th image and class label, and $N_S$ is the number of samples in the source domain. The goal of single domain generalization is to learn a domain-agnostic model with only $\mathcal{S}$ to correctly classify the images from unseen target domains. In this case, as the training objective, we use the empirical risk minimization (ERM)~\cite{vapnik1999nature} as follows:
\vspace{-0.3cm}
\begin{equation}
\underset{\phi}{\mathrm{argmin}} \frac{1}{N_S}\sum^{N_S}_{n=1} \ell(f_\phi(\mathbf{x}_n),y_n),
\label{eq:ERM}
\vspace{-0.2cm}
\end{equation}
\noindent where $f_\phi(\cdot)$ is the base network, including a feature extractor and a classifier, $\phi$ is the set of the parameters of the base network, and $l$ is a loss function measuring prediction error. Although ERM showed significant achievements on domain generalization datasets for multiple source domains~\cite{gulrajani2020search}, unfortunately, using the vanilla empirical risk minimization only with the single source domain $\mathcal{S}$ could be sub-optimal~\cite{NIPS2018_7779, fan2021adversarially} and is prone to overfitting. To derive domain-agnostic feature representations, we concentrate on generating novel styles based on random convolutions.

\subsection{Revisiting random convolutions}

In the subsection, we briefly discuss about RandConv~\cite{xu2020robust}, which forms the basis of our proposed framework. This is an image augmentation technique that applies a convolution layer with random weights on the input image, as shown in Fig.~\ref{fig:concept}(a). When RandConv is applied to an input image, it tends to modify the texture of the input image while keeping the shape of objects intact. Essentially, this layer is initialized differently for each mini-batch during training, making the network robust to different textures and removing the texture bias inherently present in CNNs. Mathematically, the convolution weights $\bm{w}\in\mathbb{R}^{k\times k \times C_{in} \times C_{out}}$ are sampled from a Gaussian distribution $N(0, \frac{1}{k^2 C_{in}})$, where $C_{in} (=\! 3)$ and $C_{out} (=\! 3)$ are the number of image channels in the input and output, and $k$ is the kernel size of the convolution layer. The kernel size $k$ is uniformly sampled from a pool $\{1,3,5,7\}$ to produce multiple-scale data augmentation. Finally, we can obtain the augmented image $\mathbf{x}'$ as $\mathbf{x}' = \mathbf{x}*\bm{w}$, where $*$ is the convolution operation. The advantage of RandConv is simple and lightweight compared to complex generators or adversarial data augmentation.

However, RandConv suffers from structural limitations. Firstly, a large-sized kernel can become an ineffective data augmentation type that interferes with learning generalizable visual representations due to excessively increased randomness and semantic distortions. In particular, this issue occurs more frequently as a larger range of pixels is affected as the kernel size increases. Figure~\ref{fig:images}(a) shows that artificial patterns can be easily created from large-sized kernels. As a result, as the kernel size increases, the generalization ability in the test domain cannot be prevented from dropping sharply, as shown in Fig.~\ref{fig:concept}(c). Secondly, RandConv lacks the inherent diversity of a single convolution operation. In this paper, we strive to address these limitations of RandConv through a progressive augmentation scheme and an advanced design that can create more diverse styles.

\section{Proposed method}

\subsection{Progressive random convolutions}

We aim to improve the style diversity while maintaining class-specific semantic information, which is a common goal of most data augmentation methods~\cite{NIPS2018_7779, zhaoNIPS20maximum, qiao2020learning, wang2021learning, li2021progressive} in domain generalization. To this end, we propose a progressive approach that repeatedly applies a convolution layer with a small kernel to the image. The structure of our method is described in Fig.~\ref{fig:concept}(b). Our progressive approach has two main properties. The first property is that pixels at the center of a theoretical receptive field have a much larger impact on output like a Gaussian, which is discussed in~\cite{luo2016understanding}. As a result, it tends to relax semantic distortions by reducing the influence of pixels away from the center. The second property is that stacking random convolution layers of the same weights can create more effective virtual domains by gradually increasing the style diversity. This approach enables more fine-grained control in image transformation. Especially, using multiple kernels with the same weight enables a more balanced and consistent application of a single transformation, whereas using different weights for each layer can result in overexposure to various transformation sets, leading to semantic distortion easily. As shown in Fig.~\ref{fig:concept}(c), our innovative change from a large kernel size to a progressive approach contributes to substantial performance improvement.

\subsection{Random convolution block}

\begin{figure}[t!]
\centering
\includegraphics[width=0.95\linewidth]{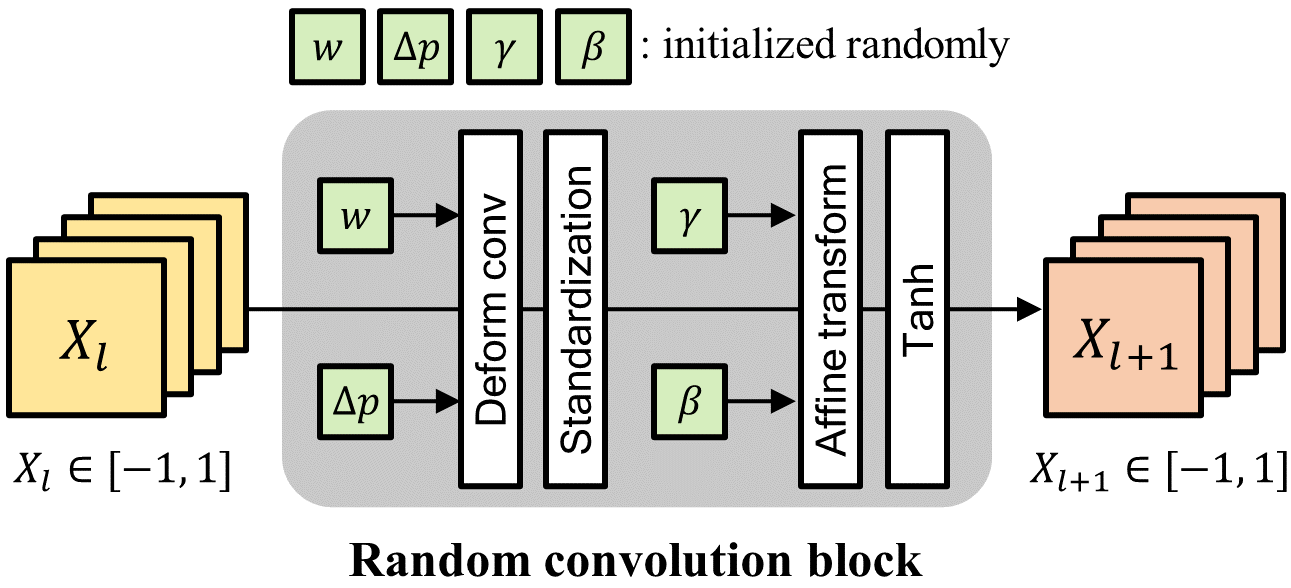}
\caption{The proposed random convolution block to support texture and contrast diversification.}
\label{fig:rand_block}
\vspace{-0.3cm}
\end{figure}

To create more diverse styles, we introduce an advanced design in which a single convolution layer is replaced with a random convolution block while preserving the progressive structure, which is depicted in Fig.~\ref{fig:rand_block}. In addition to a convolution layer, our convolution block consists of a sequence of deformable convolution, standardization, affine transform, and hyperbolic tangent function to support texture and contrast diversification. We emphasize that all parameters in the convolution block are sampled differently for each mini-batch, but are used the same in the progressive process without additional initialization as with the second property in the previous subsection.

\begin{figure}[t!]
\centering
\includegraphics[width=0.50\linewidth]{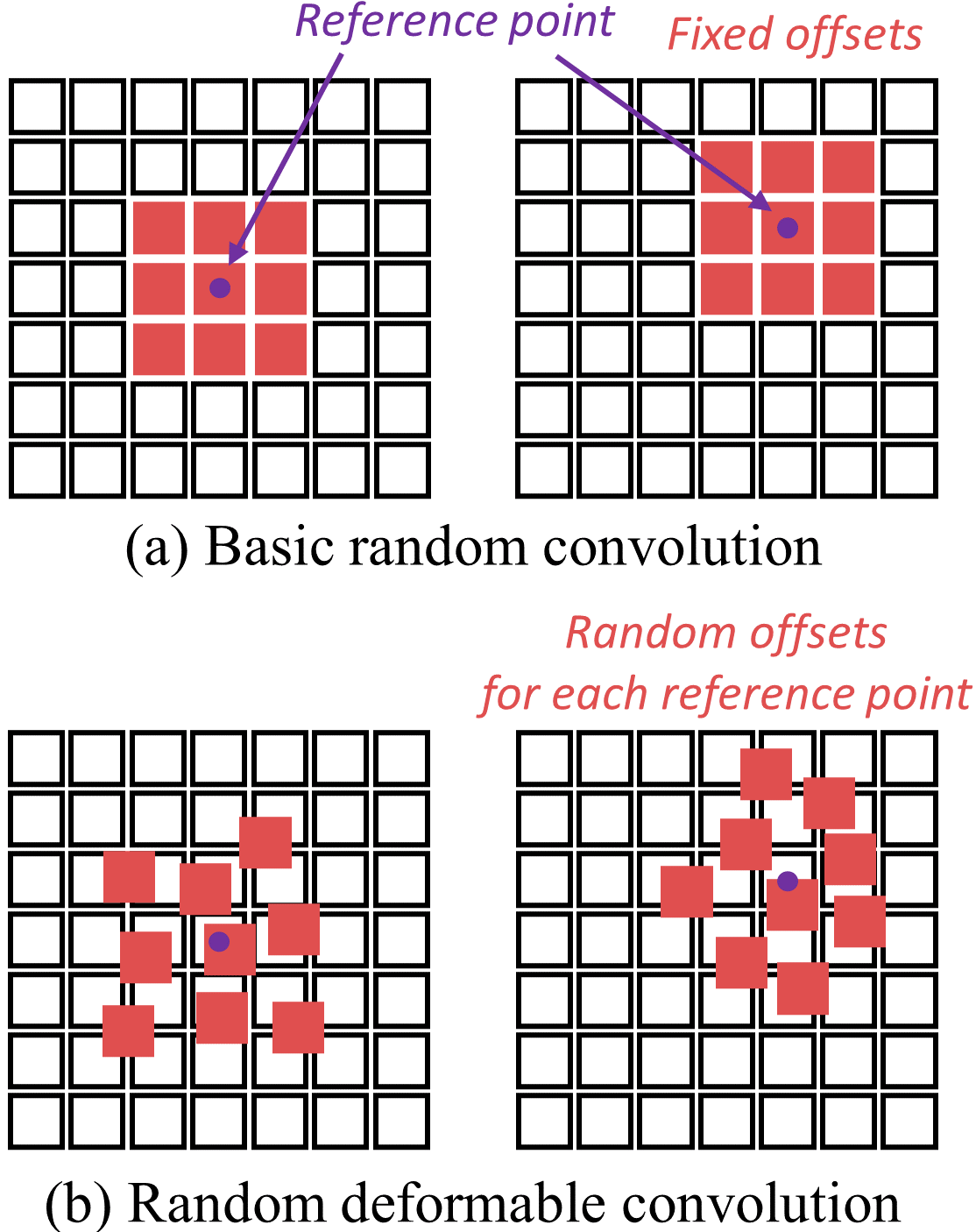}
\vspace{-0.1cm}
\caption{The difference between random convolution and random deformable convolution. In the random deformable convolution, weights are the same for each reference point, but offsets are assigned differently, which promotes the diversity of local textures.}
\label{fig:deformable_conv}
\vspace{-0.2cm}
\end{figure}

\paragraph{Texture diversification}
Given the randomized convolution weights, the conventional convolution is applied equally to each position in the image, as shown in Fig.~\ref{fig:deformable_conv}(a). We develop the random convolution operation by relaxing the constraints on the fixed grid, which can further increase the diversity of texture modifications. We adopt a deformable convolution layer~\cite{dai2017deformable} with randomly initialized offsets, which is a generalized version of the random convolution layer~\cite{xu2020robust}. We assume the 2D deformable convolution with a regular grid $\mathcal{R}$. For example, $\mathcal{R}=\{(-1,-1),(-1,0), ..., (0,1), (1,1)\}$ means a 3$\times$3 kernel with dilation 1. The operation of deformable convolution can be expressed as
\vspace{-0mm}
\begin{equation}
\mathbf{x}' [i_0, j_0]=\!\!\!\!\!\!\!\!\sum_{\substack{(i_m, j_m) \in \mathcal{R}}} \!\!\!\!\!\!\bm{w}[i_m, j_m]\cdot \mathbf{x} [i_0+i_m+\Delta i_m, j_0+j_m+\Delta j_m], 
\label{eq:deformable_conv}
\end{equation}
\vspace{-0mm}
\noindent where $i_m$, $j_m$ indicate the locations in the regular grid, $\bm{w}$ is weights of the 2D convolution kernel, and $\Delta i_m$, $\Delta j_m$ are random offsets of deformable convolution. For clarity, input and output channels are omitted. Each location $(i_0, j_0)$ on the output image $\mathbf{x}'$ is transformed by the weighted summation of weights $\bm{w}$ and the pixel values on irregular locations $(i_0+i_m+\Delta i_m, j_0+j_m+\Delta j_m)$ of the input image $\mathbf{x}$, {which is described in Fig.~\ref{fig:deformable_conv}(b).} When all offsets $\Delta i_m$, $\Delta j_m$ are allocated to zero-values, this equation is equivalent to the original random convolution layer~\cite{xu2020robust}. The offsets in deformable convolution can be considered as an extremely lightweight spatial transformer (STN)~\cite{jaderberg2015spatial}, encouraging the generation of more diverse samples that include geometric transformations. For example, when kernel size is 1, our deformable convolution covers translation as well, while the original random convolution layer changes only color.

\vspace{-0.2cm}
\paragraph{Contrast diversification}
According to \cite{xu2020robust}, assuming that the input pixel values follow the standard normal distribution, the output passed through a random convolution layer should follow the same distribution. However, when the input or convolution weights do not have an ideal distribution, the output distribution may be gradually saturated during the progressive process. To avoid this issue and diversify global style, we propose a contrast diversification method that intentionally changes the distribution of the output randomly. Given an image $\mathbf{x}'$ augmented by random convolution, we standardize the image to have zero mean and unit variance. Then, we apply an affine transformation using the randomly initialized affine parameters $\gamma_c$ and $\beta_c$ as follows:
\vspace{-1mm}
\begin{equation}
\mathbf{x}_{c}''[i, j] = \gamma_c \frac{\mathbf{x}'_{c}[i, j] - \mu_{c}}{\sqrt{\sigma^2_{c} + \epsilon}} + \beta_c,
\end{equation}
\vspace{-1mm}
\begin{equation}
\mu_{c} = \frac{\sum_{i,j} \mathbf{x}'_{c}[i, j]}{H \cdot W}, \quad \sigma^2_{c} = \frac{\sum_{i,j} \left( \mathbf{x}'_{c}[i, j] - \mu_{c} \right)^2 }{H \cdot W},
\end{equation}
\noindent where $\mu_{c}$, $\sigma^2_{c}$ are the per-channel mean and variance, respectively, and $H$, $W$ are height and width of the image. Finally, the hyperbolic tangent function transforms the image into a range between $-1$ and $1$. This entire modeling can be interpreted as a process of random gamma correction. 

% Note that contrast diversification is performed for each channel.

\vspace{-3mm}
\paragraph{Initialization}

Finally, we introduce initialization strategies. All parameters in the random convolution block are sampled from a Gaussian distribution $N(0, \sigma^2)$, as shown in Algorithm~\ref{algorithm}. However, we apply practical techniques for convolution weights and deformable offsets to control excessive texture deformation and induce realistic transformations. For convolution weights, we multiply the convolution kernel element-wise with a Gaussian filter $g[i_m,j_m] = \exp(-\frac{i_m^2+j_m^2}{2\sigma^2_g})$, where $(i_m, j_m) \in \mathcal{R}$. Specifically, $\sigma_g$ is randomly sampled per mini-batch, so the smoothing effect is different each time. This technique can be used to alleviate the problem of severely distorted object semantics when the random offset of the deformation convolution is too irregular. Next, we use a Gaussian Random Field\footnote{https://github.com/bsciolla/gaussian-random-fields} (GRF) as an initialization method for deformable offsets to induce natural geometric changes. All values of GRF basically follow a Gaussian distribution, but spatial correlation can be controlled by coefficients of the power spectrum.

\begin{algorithm}[t]
   \caption{Pro-RandConv}
   \label{algorithm}
\textbf{Input}: Source domain $\mathcal{S}=\{\mathbf{x}_n,y_n\}_{n=1}^{N_S}$ \\
\textbf{Output}: {Trained network} $f_\phi(\cdot)$ 
\begin{algorithmic}[1]
\State Initialize network parameters $\phi$
\For{$t=1$ \textbf{to} $T_{max}$} 
% \State \textbf{Base model update}: \hfill// Eq.~(\ref{eq:ce})-Eq.~(\ref{eq:base_update})
\State \textbf{Initialize a random convolution block} $\mathcal{G}$:
\State $w \sim N(0,\sigma^2_{w})$ \hfill// Convolution weights
% \State $\hat{w} \leftarrow w$ \hfill// Re-weight convolution layer
\State $\Delta p \sim N(0,\sigma^2_{\Delta})$ \hfill// Deformable offsets
\State $\gamma \sim N(0,\sigma^2_{\gamma})$ \hfill// Affine transformation (gamma)
\State $\beta \sim N(0,\sigma^2_{\beta})$ \hfill// Affine transformation (beta)
\State \textbf{Progressive augmentation}:
\State $\mathbf{X} \sim \mathcal{S}$ \hfill// Sample a mini-batch
\State $\mathbf{X}_0 \leftarrow \mathbf{X}$ \hfill// Set an initial value
% \State $L \in \{1, 2, \cdots , L_{max}\}$ \hfill// Set repetition numbers
\State $L \sim U(\{1, 2,..., L_{max}\})$ \hfill// Repetition numbers
\For{$l=1$ \textbf{to} $L$} 
\State $\mathbf{X}_{l} = \mathcal{G}(\mathbf{X}_{l-1})$ \hfill// Apply Pro-RandConv
\EndFor
\State \textbf{Training a network}:
\State $\phi \leftarrow \phi - \alpha  $ {\large\!$\nabla_{\!\phi}\!$} $\mathcal{L}_\text{task} (\mathbf{X}_{0}, \mathbf{X}_{L} ; \phi)$ \hfill// Network update
\EndFor
\end{algorithmic}
\end{algorithm}

% \vspace{-1mm}
\subsection{Training pipeline}
% \vspace{-1mm}

For each mini-batch, we initialize a random convolution block $\mathcal{G}$, and then progressively augment a set of images by the selected number of repetitions $L$. In the end, we train a network by minimizing the empirical risk in Eq.~\ref{eq:ERM} realized as the cross-entropy loss of the original and augmented images.\footnote{Besides using both original and augmented images, other strategies for choosing images to augment are explored in the supplementary Section~\ref{supp_sec3}.} Algorithm~\ref{algorithm} shows the entire training process, which is simple and easy to implement. Note that it does not require additional loss functions or complex training pipelines.

\vspace{-1mm}
\section{Experiments}
\subsection{Datasets and settings}

\textbf{Digits} dataset consists of five sub-datasets: MNIST~\cite{MNIST}, SVHN~\cite{SVHN}, MNIST-M~\cite{DANN}, SYN~\cite{DANN}, and USPS~\cite{USPS}. Each sub-dataset is regarded as a different domain containing 10-digit handwritten images in a different style. \textbf{PACS}~\cite{li2017deeper} contains four domains (Art, Cartoon, Photo, and Sketch), and each domain shares seven object categories (dog, elephant, giraffe, guitar, house, horse, and person). This dataset includes 9,991 images. \textbf{OfficeHome}~\cite{venkateswara2017deep} is an object recognition benchmark consisting of four domains (Art, Clipart, Product, and Real-World). The whole dataset consists of 15,500 images across 65 classes. \textbf{VLCS}~\cite{torralba2011unbiased} is an image classification dataset that contains 10,729 images aggregated from four datasets (PASCAL VOC 2007~\cite{everingham2007pascal}, LabelMe~\cite{russell2008labelme}, Caltech101~\cite{fei2004learning}, and Sun09~\cite{xiao2010sun}) across five shared categories (bird, car, chair, dog, and person). In addition, we extend our work to the task of \textbf{semantic segmentation}.\footnote{Segmentation results are covered in the supplementary Section~\ref{supp_sec2}.} 

% and cover the following datasets: Cityscapes~\cite{cordts2016cityscapes}, BDD-100K~\cite{yu2020bdd100k}, Mapillary~\cite{neuhold2017mapillary}, GTAV~\cite{richter2016playing}, and SYNTHIA~\cite{ros2016synthia}.

\vspace{-2mm}
\paragraph{Implementation details}
In Digits, we utilize LeNet~\cite{lecun1989backpropagation} as a base network. We train the network using SGD with batch size 64, momentum 0.9, initial learning rate 0.01, and cosine learning rate scheduling for 500 epochs. All images are resized to 32$\times$32 pixels. In PACS, OfficeHome, and VLCS, we employ AlexNet~\cite{krizhevsky2012imagenet}, ResNet18~\cite{he2016deep}, and ResNet50~\cite{he2016deep}, respectively. We train the network using SGD with batch size 64, momentum 0.9, initial learning rate 0.001, and cosine learning rate scheduling for 50 epochs and 150 epochs in single DG and multi DG experiments, respectively. All images are resized to 224$\times$224. For a fair comparison, we follow the experimental protocol as in~\cite{zhou2021domain, liu2022geometric}. We used in-domain validation sets to choose the best model from multiple saved checkpoints during training, as Training-Domain Validation in DomainBed~\cite{gulrajani2020search}.

% For all experiments, we choose the model with the highest performance in the validation set.
Unlike RandConv~\cite{xu2020robust}, we fix the kernel size $k$ of the convolution to 3 and set the maximum number of repetitions $L_{max}$ to 10, which means that for each mini-batch, the convolution block is recursively stacked a different number of times chosen between 1 and 10. For initialization, we set $\sigma_{w}$ to $1/\sqrt{k^2 C_{in}}$, known as He-initialization~\cite{he2015delving}. Furthermore, we re-weight the convolution weights once more by Gaussian smoothing $g[i_m,j_m] = \exp(-\frac{i_m^2+j_m^2}{2\sigma^2_g})$, where the scale is sampled from $\sigma_g \sim U(\epsilon, 1)$, where $\epsilon$ indicates a small value. We sample the deformable offsets from $N(0, \sigma^2_\Delta)$, where the scale is sampled from uniform distribution as $\sigma_\Delta \sim U(\epsilon, b_\Delta)$. We use $b_\Delta$ of 0.2 for Digits and 0.5 for other datasets because the deformation scale is affected by the size of the image. In contrast diversification, we set both $\sigma_{\gamma}$ and $\sigma_{\beta}$ to 0.5. Please refer to the supplementary Section~\ref{supp_sec6} for more details.

\subsection{Evaluation of single domain generalization}

\begin{table}[t]
\caption{Single domain generalization accuracy (\%) trained on MNIST. Each column title indicates the target domain, and the numerical values represent its performance. LeNet is used for training. * denotes reproduced results. The two versions of the progressive approach indicate whether the random convolution layers are stacked with the same or different weights. Pro-RandConv is the final version with the random convolution blocks added.}
\begin{center}
\vspace{-5mm}
\scalebox{0.8}{
\begin{tabular}{l|cccc|c}
\hline
Methods & SVHN & MNIST-M & SYN & USPS & Avg. \\
\hline
CCSA \cite{CCSA}           & 25.89 & 49.29 & 37.31 & 83.72 & 49.05 \\
d-SNE \cite{d-sne}          & 26.22 & 50.98 & 37.83 & \textbf{93.16} & 52.05 \\
JiGen \cite{carlucci2019domain}           & 33.80 & 57.80 & 43.79 & 77.15 & 53.14 \\
ADA \cite{NIPS2018_7779}            & 35.51 & 60.41 & 45.32 & 77.26 & 54.62 \\
M-ADA \cite{qiao2020learning}           & 42.55 & 67.94 & 48.95 & 78.53 & 59.49 \\
ME-ADA \cite{zhaoNIPS20maximum}         & 42.56 & 63.27 & 50.39 & 81.04 & 59.32 \\
L2D \cite{wang2021learning}           & 62.86 & 87.30 & 63.72 & 83.97 & 74.46 \\
PDEN \cite{li2021progressive}           & 62.21 & 82.20 & 69.39 & 85.26 & 74.77 \\
MetaCNN \cite{wan2022meta}        & \textbf{66.50} & \textbf{88.27} & \textbf{70.66} & 89.64 & \textbf{78.76} \\
\hline
Baseline (ERM) & 32.52 & 54.92 & 42.34 & 78.21 & 52.00\\
RandConv*~\cite{xu2020robust}        & 61.66 & \textbf{84.53} & 67.87 & 85.31 & 74.84 \\
Progressive (Diff) & 60.73 & 78.47 & 71.46 & 88.20 & 74.72 \\
Progressive (Same) & 65.67 & 76.26 & 77.13 & \textbf{93.98} & 78.26 \\
\textbf{Pro-RandConv}    & \textbf{69.67} & 82.30 & \textbf{79.77} & 93.67 & \textbf{81.35} \\
\hline
\end{tabular}
}
\vspace{-2mm}
\end{center}
\label{table:single-digits}
\end{table}

\begin{table}[t]
\caption{Single domain generalization accuracy (\%) on PACS. Each column title indicates the source domain, and the numerical values represent the average performance in the target domains. ResNet18 is used for training. * denotes reproduced results.}
\begin{center}
\vspace{-5mm}
\scalebox{0.8}{
\begin{tabular}{l|cccc|c}
\hline
Methods & Art & Cartoon & Photo & Sketch & Avg. \\
\hline
JiGen \cite{carlucci2019domain}          & 67.70 & 72.23 & 41.70 & 36.83 & 54.60 \\
ADA \cite{NIPS2018_7779}            & 72.43 & 71.97 & 44.63 & 45.73 & 58.70 \\
SagNet \cite{nam2021reducing}          & 73.20 & 75.67 & 48.53 & 50.07 & 61.90 \\
GeoTexAug \cite{liu2022geometric}             & 72.07 & \textbf{78.70} & 49.07 & \textbf{59.97} & 65.00 \\
L2D \cite{wang2021learning}            & \textbf{76.91} & 77.88 & \textbf{52.29} & 53.66 & \textbf{65.18} \\
\hline
Baseline (ERM) & 74.64 & 73.36 & 56.31 & 48.27 & 63.15\\
RandConv*~\cite{xu2020robust}       & 76.93 & 76.47 & 62.46 & 54.13 & 67.50 \\
Progressive (Diff)  & 75.46 & 75.39 & 60.02 & 55.02 & 66.47 \\
Progressive (Same)  & 76.81 & 78.27 & 62.38 & 56.08 & 68.39 \\
\textbf{Pro-RandConv}    & \textbf{76.98} & \textbf{78.54} & \textbf{62.89} & \textbf{57.11} & \textbf{68.88} \\
\hline
\end{tabular}
}
\vspace{-2mm}
\end{center}
\label{table:single-pacs}
\end{table}

\paragraph{Results on Digits}
In Table~\ref{table:single-digits}, we compare our proposed method in the Digits benchmark. We use MNIST as the source domain and train the model with the first 10,000 samples from 60,000 training images. The remaining four sub-datasets are employed as the target domains. We observed that our method has the following properties: 1) Our method outperforms data augmentation methods\footnote{Policy-based image augmentation methods (\ie AutoAugment and RandAugment) are discussed in the supplementary Section~\ref{supp_sec5}.} based on adversarial learning (ADA~\cite{NIPS2018_7779}, ME-ADA~\cite{zhaoNIPS20maximum}, M-ADA~\cite{qiao2020learning}, L2D~\cite{wang2021learning}, and PDEN~\cite{li2021progressive}) only by image augmentation without a complicated training pipeline or multiple objective functions. 2) Compared with the PDEN method that progressively expands the domain using a learning-based generator, our method shows that the domain can be expanded through a simple random convolution block without learning. 3) The SVHN, SYN, and USPS datasets have a large domain gap with MNIST in terms of font shapes. Our Pro-RandConv method produces significantly improved recognition performance compared to RandConv~\cite{xu2020robust} on these domains due to the progressive approach.

\vspace{-2mm}
\paragraph{Results on PACS} 
We also compare our method on the challenging PACS benchmark, as shown in Table~\ref{table:single-pacs}. Our method outperforms all of the competitors by a large margin. It is noteworthy that generalizable representations can be learned with a simple image augmentation by random convolutions, unlike a geometric and textural augmentation method~\cite{liu2022geometric} that borrows elaborate sub-modules (\eg an arbitrary neural artistic stylization network~\cite{ghiasi2017exploring} and a geometric warping network~\cite{liu2021learning}) to generate novel styles. Compared to the existing RandConv~\cite{xu2020robust}, the performance improvement on cartoon and sketch domains is remarkable.

\subsection{Evaluation of multi domain generalization}
To further validate the performance of our method, we conducted multi DG experiments on PACS. In multi DG experiments, we use a leave-one-domain-out protocol, where we choose one domain as the test domain and use the remaining three as source domains for model training. Table~\ref{table:multi-pacs} shows that our method outperforms other DG methods except for Fourier Augmented Co-Teacher (FACT)~\cite{xu2021fourier} and Geometric and Textural Augmentation (GeoTexAug)~\cite{liu2022geometric}. It is meaningful that the performance is much more competitive than the traditional RandConv~\cite{xu2020robust}. This proves that the progressive approach and diversification techniques are highly effective.

\begin{table}[t]
\caption{Multi domain generalization accuracy (\%) on PACS. Each column title is the target domain, and the numerical values represent its performance. The remaining three domains that are not the target domain are used for training. * denotes reproduced results.}
\begin{center}
\vspace{-5mm}
\scalebox{0.79}{
\begin{tabular}{l|cccc|c}
\hline
Methods & Art & Cartoon & Photo & Sketch & Avg. \\
\hline
\multicolumn{6}{c}{ResNet-18} \\
\hline
Jigen \cite{carlucci2019domain}          & 79.42 & 75.25 & 96.03 & 71.35 & 80.51 \\
MASF \cite{dou2019domain}           & 80.29 & 77.17 & 94.99 & 71.68 & 81.03 \\
ADA \cite{NIPS2018_7779}            & 78.32 & 77.65 & 95.61 & 74.21 & 81.44 \\
Epi-FCR \cite{li2019episodic}        & 82.10 & 77.00 & 93.90 & 73.00 & 81.50 \\
MetaReg \cite{balaji2018metareg}         & 83.70 & 77.20 & 95.50 & 70.30 & 81.70 \\
ME-ADA \cite{zhaoNIPS20maximum}          & 78.61 & 78.65 & 95.57 & 75.59 & 82.10 \\
EISNet \cite{wang2020learning}         & 81.89 & 76.44 & 95.93 & 74.33 & 82.15 \\
InfoDrop \cite{shi2020informative}       & 80.27 & 76.54 & 96.11 & 76.38 & 82.33 \\
L2A-OT \cite{zhou2020learning}         & 83.30 & 78.20 & \textbf{96.20} & 73.60 & 82.80 \\
DDAIG \cite{zhou2020deep}          & 84.20 & 78.10 & 95.30 & 74.70 & 83.10 \\
SagNet \cite{nam2021reducing}          & 83.58 & 77.66 & 95.47 & 76.30 & 83.25 \\
MixStyle \cite{zhou2021domain}        & 84.10 & 78.80 & 96.10 & 75.90 & 83.70 \\
L2D \cite{wang2021learning}            & 81.44 & \textbf{79.56} & 95.51 & \textbf{80.58} & 84.27 \\
FACT \cite{xu2021fourier}           & \textbf{85.37} & 78.38 & 95.15 & 79.15 & \textbf{84.51} \\
\hline
Baseline (ERM)       & 81.54 & 80.06 & 95.80 & 68.40 & 81.45 \\
RandConv*~\cite{xu2020robust}       & 80.15 & 78.04 & 93.65 & \textbf{77.88} & 82.43 \\
\textbf{Pro-RandConv}    & \textbf{83.15} & \textbf{81.07} & \textbf{96.24} & 76.71 & \textbf{84.29} \\
\hline
\multicolumn{6}{c}{ResNet-50} \\
\hline
MASF \cite{dou2019domain}           & 82.89 & 80.49 & 95.01 & 72.29 & 82.67 \\
MetaReg \cite{balaji2018metareg}       & 87.20 & 79.20 & 97.60 & 70.30 & 83.60 \\
EISNet \cite{wang2020learning}         & 86.64 & 81.53 & 97.11 & 78.07 & 85.84 \\
FACT \cite{xu2021fourier}             & 89.63 & 81.77 & 96.75 & 84.46 & 88.15 \\
GeoTexAug \cite{liu2022geometric}            & \textbf{89.98} & \textbf{83.84} & \textbf{98.10} & \textbf{84.75} & \textbf{89.17} \\
\hline
Baseline (ERM)       & 87.15 & 83.82 & 97.77 & 73.71 & 85.61 \\
RandConv*~\cite{xu2020robust}       & 86.13 & 82.00 & 96.65 & 81.72 & 86.63 \\
\textbf{Pro-RandConv}    & \textbf{89.28} & \textbf{84.13} & \textbf{97.83} & \textbf{81.85} & \textbf{88.27} \\
\hline
\end{tabular}
}
\vspace{-3mm}
\end{center}
\label{table:multi-pacs}
\end{table}

\subsection{Discussion}

\begin{table}[t]
\caption{Performance comparison of RandConv (RC) and our Pro-RandConv (P-RC) on different datasets and models. Each value represents the average domain generalization accuracy (\%).}
\begin{center}
\vspace{-5mm}
\scalebox{0.73}{
\begin{tabular}{c|c|ccc|ccc}
\hline
\multirow{2}{*}{Dataset} & \multirow{2}{*}{Model} &  \multicolumn{3}{c|}{Single DG}  &  \multicolumn{3}{c}{Multi DG} \\
\cline{3-8} 
 &  & RC~\cite{xu2020robust} & \textbf{P-RC} & $\Delta$ & RC~\cite{xu2020robust} & \textbf{P-RC} & $\Delta$ \\
\hline
        & Alex & 60.71 & \textbf{62.59} & $+$1.88 & 72.49 & \textbf{74.91} & $+$2.42 \\
PACS    & Res18 & 67.50 & \textbf{68.88} & $+$1.38 & 82.43 & \textbf{84.29} & $+$1.86 \\
        & Res50 & 72.33 & \textbf{73.26} & $+$0.93 & 86.63 & \textbf{88.27} & $+$1.64 \\
\hline
\multirowcell{3}{Office\\Home}   & Alex & 42.04 & \textbf{43.25} & $+$1.21 & 53.92 & \textbf{55.71} & $+$1.79 \\
  & Res18 & 50.61 & \textbf{51.32} & $+$0.71 & 63.08 & \textbf{64.59} & $+$1.51 \\
            & Res50 & 58.57 & \textbf{59.20} & $+$0.63 & 68.83 & \textbf{69.89} & $+$1.06 \\
\hline
        & Alex & 56.80 & \textbf{60.70} & $+$3.90 & 69.74 & \textbf{71.30} & $+$1.56 \\
VLCS    & Res18 & 53.05 & \textbf{53.35} & $+$0.30 & 67.83 & \textbf{69.55} & $+$1.72 \\
        & Res50 & 53.57 & \textbf{54.23} & $+$0.66 & 70.26 & \textbf{71.68} & $+$1.42 \\
\hline
\end{tabular}
}
\vspace{-4mm}
\end{center}
\label{table:comparison}
\end{table}

\begin{figure}[t!]
\centering
\includegraphics[width=\linewidth]{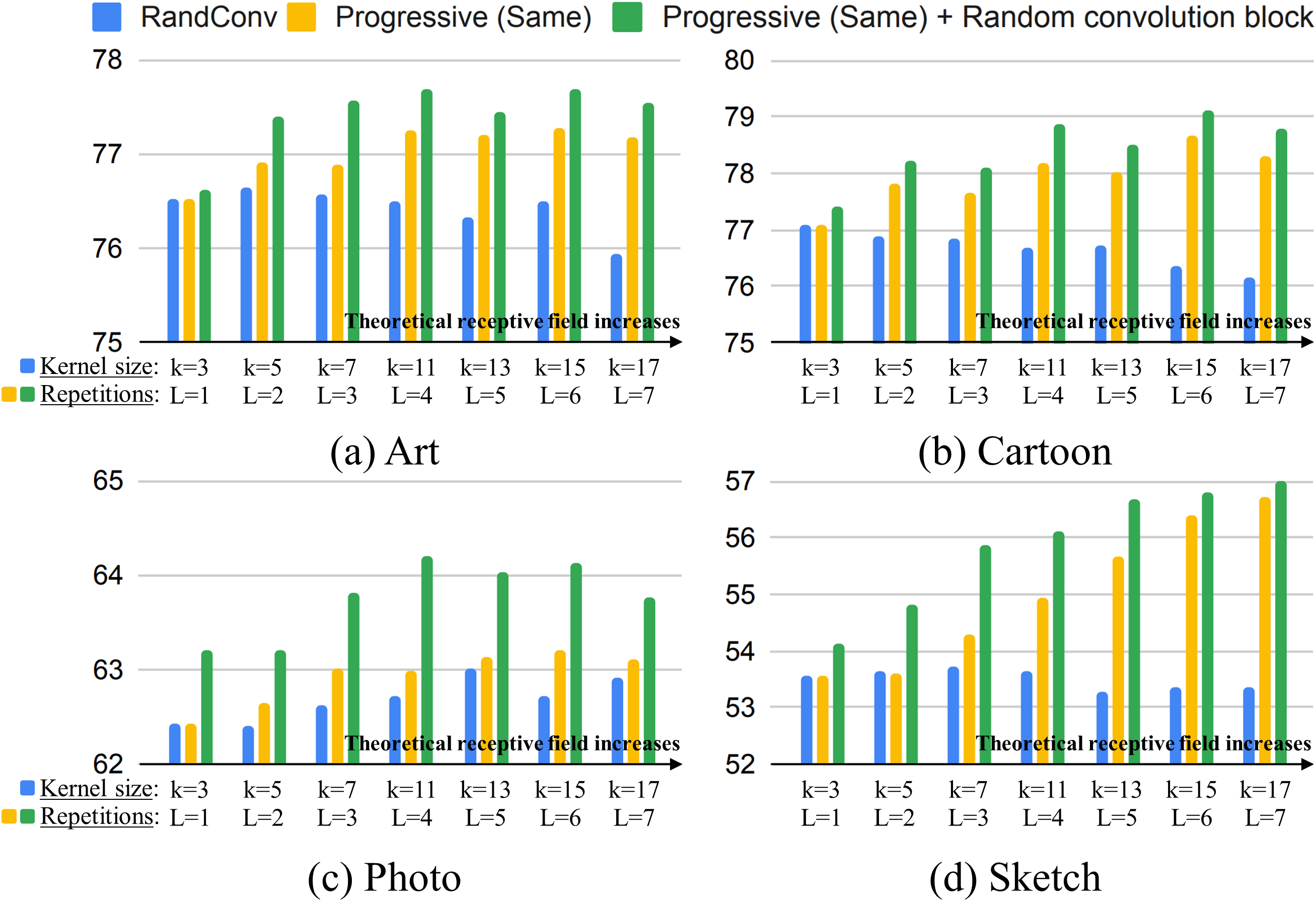}
\vspace{-0.7cm}
\caption{Comparative analysis of changing the kernel size ($k$) of RandConv and changing the number of repetitions ($L$) of Pro-RandConv in the single domain generalization setting.}
\label{fig:PACS_plot}
\vspace{-0.25cm}
\end{figure}

\begin{figure}[t!]
\centering
\includegraphics[width=0.75\linewidth]{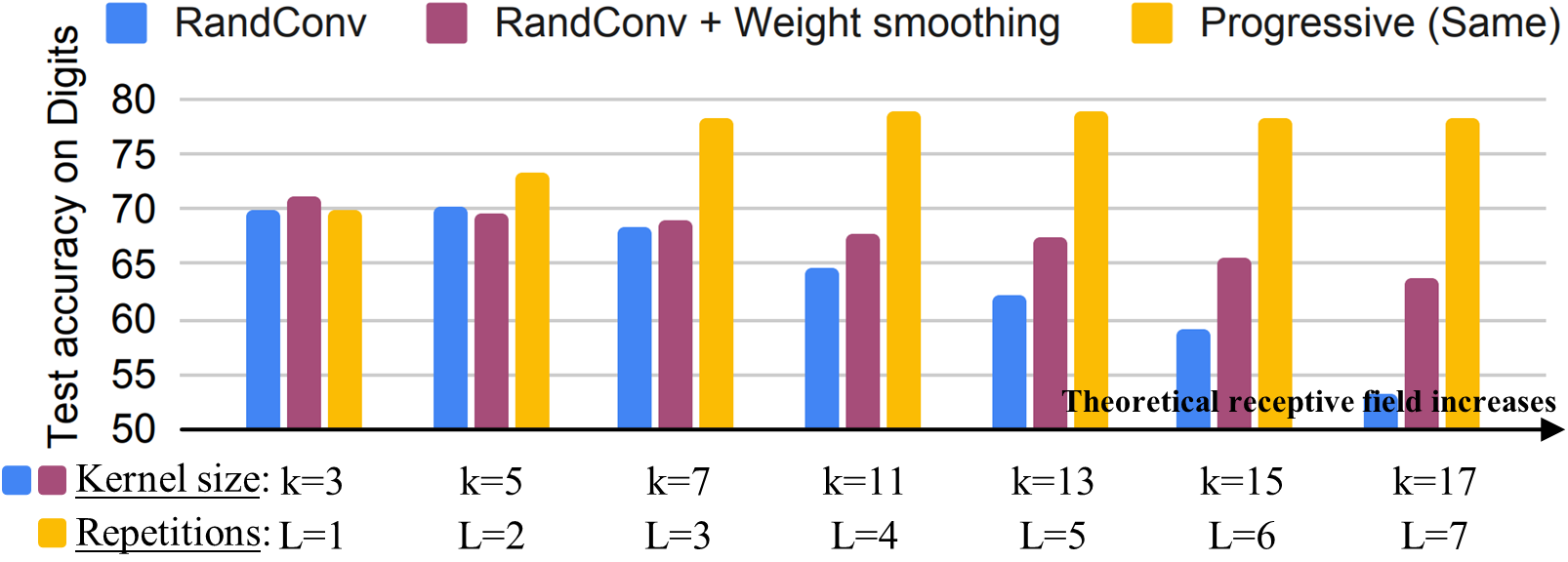}
\vspace{-0.2cm}
\caption{Analysis of applying Gaussian smoothing to a convolution kernel of RandConv in the single domain generalization setting.}
\label{fig:smoothing}
\vspace{-2.5mm}
\end{figure}

\paragraph{Comparative analysis with RandConv}
In Table.~\ref{table:comparison}, we compare RandConv~\cite{xu2020robust} and Pro-RandConv using AlexNet~\cite{krizhevsky2012imagenet}, ResNet18~\cite{he2016deep} and ResNet50~\cite{he2016deep} models on PACS, OfficeHome, and VLCS datasets. We emphasize that the proposed Pro-RandConv outperforms RandConv in all datasets and all models. In addition, this performance improvement is more pronounced in multi DG experiments. Through these results, we show that our progressive approach using a random convolution block significantly contributes to the improvement of generalization capability by generating more effective virtual domains.

In Fig.~\ref{fig:PACS_plot}, we compare RandConv with our progressive models on PACS. Since the image size of PACS is larger than that of digits, the performance does not drop sharply as the kernel size of RandConv increases. However, the ability of RandConv to create novel styles for learning domain generalizable representations is significantly different from that of the progressive model with the same weight (\textbf{\textcolor{blue}{blue}}$\longleftrightarrow$\textbf{\textcolor{yellow}{yellow}}). In other words, using a large-sized kernel is a less effective data augmentation technique compared to the repeated use of a small-sized kernel. This can be considered an inherent problem of the single convolution operation in RandConv. In addition, the performance gap between the presence and absence of random convolution blocks is the largest in the photo domain, which is a meaningful observation that diversification techniques work well on realistic images (\textbf{\textcolor{yellow}{yellow}}$\longleftrightarrow$\textbf{\textcolor{green}{green}}).

\begin{figure*}[t!]
\centering
\vspace{-3mm}
\includegraphics[width=1.00\linewidth]{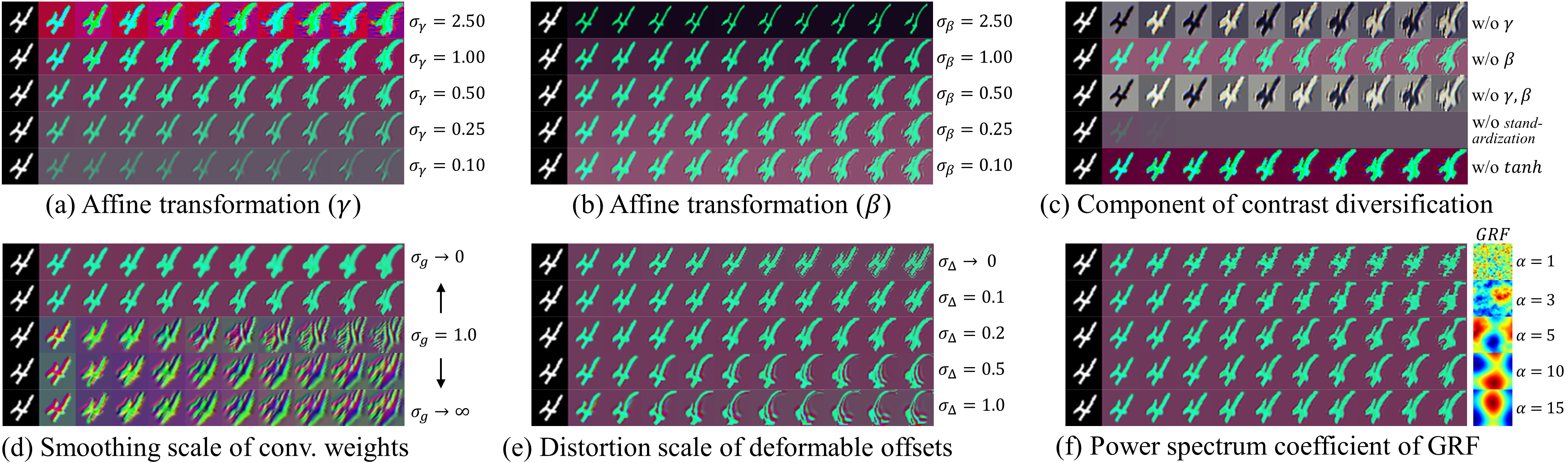}
\vspace{-5mm}
\caption{Visual analysis of individual components and weight initialization in the proposed random convolution block. (a)-(c) are related to contrast diversification and (d)-(e) are related to texture diversification.}
\vspace{-2mm}
\label{fig:diversification}
\end{figure*}

We hypothesized that the progressive approach with equal weights works well for two reasons: The first is to mitigate semantic distortion by reducing the influence of pixels away from the center, and the second is to create effective virtual domains by gradually increasing the distortion magnitude to the central pixels. To indirectly validate the first assumption, we applied Gaussian smoothing to a convolution kernel of RandConv. Figure~\ref{fig:smoothing} shows that the performance decreases less compared to the conventional RandConv as the size of the kernel increases. In other words, the Gaussian-like effective receptive field~\cite{luo2016understanding} can help mitigate semantic distortion. Beyond that, the reason why the progressive approach performs much better is that even if the effective receptive field is similar, our progressive approach enables more fine-grained control in image transformation than a single layer with randomly initialized.

\vspace{-3mm}
\paragraph{Analysis on texture and contrast diversification}
We introduce the effects of contrast diversification and texture diversification through qualitative results. First, contrast diversification consists of affine transformation, standardization, and hyperbolic tangent function, as shown in Fig.~\ref{fig:rand_block}. Figure~\ref{fig:diversification}(a) and (b) show that gamma correction is differently performed according to $\sigma_\gamma$ and $\sigma_\beta$ in affine transformation. If the standard deviation for affine transformation is too large or too small, erroneous distortion or saturation can occur in the image. In Fig.~\ref{fig:diversification}(c), we check the qualitative results by disabling individual components one by one. If there is no affine transformation, the contrast diversity is limited because the texture or color of the image is just determined by the input distribution and the convolution weight. When a standardization layer is removed, a saturation problem occurs. Finally, the hyperbolic tangent function reduces the image saturation and stabilizes the distribution.

Next, Fig.~\ref{fig:diversification}(d) shows the result of applying Gaussian smoothing to the 3$\times$3 convolution kernel of Pro-RandConv. When $\sigma_g$ approaches 0. it simulates a 1$\times$1 kernel. When $\sigma_g$ approaches infinity, it simulates a normal 3$\times$3 kernel ignoring smoothing. This implies that we can suppress excessive deformation of texture through this smoothing function. We sample it from a uniform distribution as $\sigma_g \sim U(\epsilon,1)$ to reduce the sensitivity of the hyperparameter. Figure~\ref{fig:diversification}(e) shows the distortion scale of deformable offsets. This parameter is related to image size, so we use different values considering the image size. Figure~\ref{fig:diversification}(f) shows the coefficients of the power spectrum in a Gaussian Random Field. The lower the coefficient, the more similar the shape of the white noise. We use reasonably high values to induce natural transformations. Please refer to supplementary Sections \ref{supp_sec4} and \ref{supp_sec6} for component analysis of Pro-RandConv and hyperparameter selection.

\begin{table}[t]
\vspace{+0.1cm}
\caption{Efficiency statistics evaluated on RTX A5000 (64/batch).}
\begin{center}
\vspace{-0.4cm}
\scalebox{0.66}{
\begin{tabular}{c|c|c|c|c|c|c}
\hline
\multicolumn{2}{c|}{Digits | Size:[3,32,32]}  & \multicolumn{2}{c|}{Training}  & \multicolumn{2}{c|}{Inference} & Acc \\
\cline{1-6}
Methods & Versions &Memory&Time&MACs&Time&(\%)\\
\hline
ERM & -                         & \cred{2.11GB} & \cred{3.30ms} & \multirowcell{6}{28.77M} & \multirowcell{6}{0.53ms} & \cblue{52.00} \\ 
\cline{1-4}\cline{7-7}
PDEN~\cite{li2021progressive} & -                        & \cblue{2.39GB} & \cblue{31.8ms} &  & & 73.89  \\ 
\cline{1-4}\cline{7-7}
\multirowcell{2}{RandConv \\ \cite{xu2020robust}} & w/o KLD  & 2.11GB & 4.67ms &  &  & 73.74  \\  
 & w/ KLD (default)             & 2.22GB & 13.0ms &  &  & 74.84  \\ 
\cline{1-4}\cline{7-7}
\multirowcell{2}{\textbf{Ours}} & Progressive (Same) & 2.17GB & 6.47ms & & & 78.26 \\ 
 & + RC Block                   & 2.17GB & 12.0ms &  &  & \cred{81.35}  \\ 
\hline
\end{tabular}
}
\vspace{-0.46cm}
\end{center}
\label{table:Digits}
\end{table}

\paragraph{Computational complexity and inference time}

In the training process, the computational complexity (MACs) is increased by 0.45M in RandConv and 0.92M in Pro-RandConv, representing a relatively small increase of 2-3\% compared to the base model (28.77M MACs). And the training time also increases, primarily caused by the Kullback-Leibler (KL) divergence loss in RandConv and the initialization process of the deformable operator in Pro-RandConv. However, we note that the training efficiency of Pro-RandConv is comparable to that of competitors (RandConv and PDEN~\cite{li2021progressive}), as shown in Table~\ref{table:Digits}. Moreover, image augmentation is not utilized during inference, so it does not affect complexity or time in inference.

\section{Conclusion}

We proposed Pro-RandConv as an effective image augmentation technique to enhance generalization performance in domain generalization tasks. Our approach involves progressively applying small-scale random convolutions instead of a single large-scale convolution. Furthermore, we allow deformable offsets and affine transformations to promote texture and contrast diversity. We evaluated the effectiveness of our method on both single and multi-domain generalization benchmarks and found that it outperformed recent competitors significantly. We also conducted a comprehensive analysis to show that Pro-RandConv produces qualitative and quantitative improvements over the RandConv method. In the future, we look forward to seeing its impact on improving generalization performance in real-world scenarios.

% We believe that Pro-RandConv has the potential to be widely used in various applications and

% \newpage

%%%%%%%%% REFERENCES
{\small
\bibliographystyle{ieee_fullname}
\bibliography{cvpr}
}

\clearpage

\appendix
  \renewcommand\thesection{\arabic{section}}
  % \renewcommand\thesection{\Alph{section}}

%\Huge
%\huge
%\LARGE
%\Large

\noindent{\LARGE \textbf{Supplementary Materials}}
% \vspace{5mm}

% \makeatletter
% \g@addto@macro\@maketitle{
  \begin{figure*}[t]
  \centering
  \vspace{-0.7cm}
  \includegraphics[width=\linewidth]{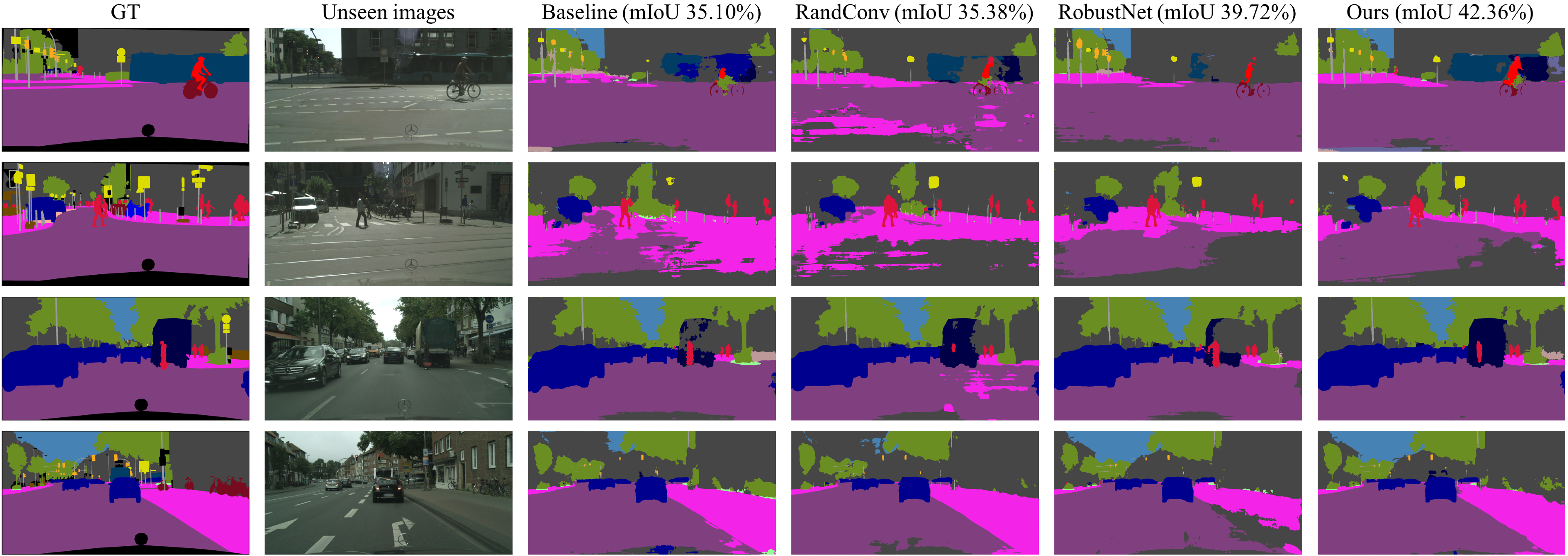}
  \vspace{-0.5cm}
  \caption{Segmentation results for unseen domain images. All models are trained on the GTAV~\cite{richter2016playing} train set and validated on the Cityscapes~\cite{cordts2016cityscapes} validation set. DeepLabV3+ is adopted as a baseline. Our method outperforms the baseline, RandConv~\cite{xu2020robust}, and RobustNet~\cite{choi2021robustnet} methods.}
  \label{fig:segmentation}
  \end{figure*}
% }
% \makeatother
% \maketitle

\section{Reproducibility}

We have provided implementation details and pseudocode in the main paper for reproducibility. Note that all the experiments have been performed eight times and averaged.

\section{Domain generalized semantic segmentation}
\label{supp_sec2}

To show the applicability of Pro-RandConv, we conducted semantic segmentation experiments in addition to the object recognition experiments provided in the main paper. We use the experimental protocol used in RobustNet~\cite{choi2021robustnet} for a fair comparison. We adopt a DeepLabV3+~\cite{chen2018encoder} architecture with ResNet50~\cite{he2016deep} as a baseline. We use the GTAV~\cite{richter2016playing} dataset as the training domain and measure the generalization capability on the Cityscapes~\cite{cordts2016cityscapes}, BDD-100K~\cite{yu2020bdd100k}, SYNTHIA~\cite{ros2016synthia}, and Mapillary~\cite{neuhold2017mapillary} datasets. Mean Intersection over Union (mIoU) is used to quantitatively evaluate semantic segmentation performance. We use a batch size of 8 for experiments on a single GPU, which is different from the experimental protocol in \cite{choi2021robustnet} using a batch size of 16 for GTAV. Except for the batch size, all environments are the same as the official experimental protocol, refer to \cite{choi2021robustnet} for more details. In our augmentation settings, we use all of the same hyperparameters for object recognition without additional tuning. We also randomly select only half of the images for each batch and perform augmentation.

Tabel~\ref{table:segmentation} shows a comparison of generalization performance in semantic segmentation. To prove the superiority of Pro-RandConv, we compare the performance not only with RandConv~\cite{xu2020robust} but also with RobustNet~\cite{choi2021robustnet}, a domain generalization method for semantic segmentation. Besides that, we compare the performance with various competitors (\eg Switchable Whitening (SW)~\cite{pan2019switchable}, IBN-Net~\cite{pan2018two}, and IterNorm~\cite{huang2019iterative}) provided by \cite{choi2021robustnet}. Our method outperforms all of the competitors including RandConv and RobustNet by a big margin. In particular, we note that our method shows a great performance improvement on real-world datasets (\ie Cityscapes, BDD-100K, and Mapillary). We also provide experimental results with various versions to observe the importance of each component. All components except the deformable offset improve the generalization performance, which can be interpreted as the geometrical change of the object shape from the deformable offset causing a negative effect on the pixel-level classification. We expect to get better generalization performance if we change the ground truth to accommodate geometric changes. Figure~\ref{fig:segmentation} describes the semantic segmentation results on Cityscapes. Ours-C version of the model with removed deformable offsets is used for visualization.

\begin{table}[t]
\caption{Performance comparison of mIoU (higher is better) (\%). The models are trained on the train set of GTAV (G), and evaluated on Cityscapes (C), BDD-100K (B), SYNTHIA (S), and Mapillary (M) validation sets. ResNet-50 is used with an output stride of 16.  DeepLabV3+ is adopted as a baseline. * denotes reproduced results. Ours-A is a version with only a progressive approach, Ours-B is a version with a progressive approach and contrast diversification, Ours-C is a version with a progressive approach, contrast diversification, and Gaussian smoothing, and Ours-D is a version with all components applied.}
\begin{center}
\vspace{-5mm}
\scalebox{0.80}{
\begin{tabular}{l|cccc|c}
\hline
Methods & C & B & S & M & Avg. \\
\hline
Baseline~\cite{choi2021robustnet}   & 28.95 & 25.14 & 26.23 & 28.18 & 27.13 \\
SW~\cite{pan2019switchable}         & 29.91 & 27.48 & 27.61 & 29.71 & 28.68 \\ 
IterNorm~\cite{huang2019iterative}  & 31.81 & 32.70 & 27.07 & 33.88 & 31.37 \\ 
IBN-Net~\cite{pan2018two}           & 33.85 & 32.30 & 27.90 & 37.75 & 32.95 \\
RobustNet~\cite{choi2021robustnet}  & 36.58 & 35.20 & \textbf{28.30} & 40.33 & 35.11 \\ 
\hline
Baseline*                           & 35.10 & 27.18 & 26.71 & 30.63 & 29.91 \\
RandConv*~\cite{xu2020robust}   & 35.38 & 30.92 & 24.45 & 32.43 & 30.80 \\
RobustNet*~\cite{choi2021robustnet} & 39.72 & 35.61 & 26.87 & 39.50 & 35.43 \\ 
\hline
Ours-A               & 39.53 & 34.14 & 26.30 & 36.74 & 34.18 \\
Ours-B               & 41.60 & 34.95 & 26.18 & 41.31 & 36.01 \\
\textbf{Ours-C}               & \textbf{42.36} & \textbf{37.03} & 25.52 & \textbf{41.63} & \textbf{36.64} \\
Ours-D               & 40.48 & 36.68 & 26.82 & 40.76 & 36.19 \\
\hline
\end{tabular}
}
\vspace{-3mm}
\end{center}
\label{table:segmentation}
\end{table}

\begin{table}[t]
\caption{Strategies for selecting training images in the single domain generalization setting on Digits and PACS in terms of accuracy (\%). LeNet and ResNet18 are used for training on Digits and PACS, respectively. RC* denotes the reproduced results of RandConv. $\mathbf{X}_{0}$ and $\mathbf{X}_{L}$ indicate original images and augmented images passing through $L$-layers. $L$ is sampled as $L \sim U(1, L_{max}=10)$ for each mini-batch, respectively. $P_{r}(\mathbf{X}_{0}, \mathbf{X}_{L})$ means an instance-level augmentation strategy, where $r$ is the data fraction of the original images. The larger $r$, the higher the proportion of the original images in the mini-batch.}
\begin{center}
\vspace{-5mm}
\scalebox{0.73}{
\begin{tabular}{c|c|cc|cc}
\hline
\multirowcell{3}{Methods} & \multirowcell{3}{Selection\\strategies}  & \multicolumn{2}{c|}{Digits} & \multicolumn{2}{c}{PACS} \\
\cline{3-6}
 & & \multirowcell{2}{In-\\domain} & \multirowcell{2}{Out-of-\\domain} & \multirowcell{2}{In-\\domain} & \multirowcell{2}{Out-of-\\domain}  \\
&&&&\\
\hline
RC*~\cite{xu2020robust} & $\mathbf{X}_{0}$ or $\mathbf{X}_{1}$ & 98.90 & 74.84 & 92.75 & 67.50 \\ 
\hline
\multirowcell{3}{Ours\\(batch)} & only $\mathbf{X}_{0}$ (baseline) & 98.64 & 52.00 & 95.37 & 63.15 \\ 
 & only $\mathbf{X}_{L}$ & 99.25 & 80.99 & 94.66 & 68.10 \\ 
 & $\mathbf{X}_{0}$ or $\mathbf{X}_{L}$ & 99.25 & 81.08 & 95.59 & 67.65 \\ 
\hline
\multirowcell{4}{Ours\\(instance)} & $P_{r=0.25}(\mathbf{X}_{0}$,$\mathbf{X}_{L})$ & 99.28 & 81.20 & 95.18 & 68.43 \\ 
 & $P_{r=0.50}(\mathbf{X}_{0}$,$\mathbf{X}_{L})$ & \textbf{99.31} & 81.13 & 95.65 & 68.20 \\ 
 & $P_{r=0.75}(\mathbf{X}_{0}$,$\mathbf{X}_{L})$& 99.25 & 80.22 & 95.73 & 67.26 \\ 
 & $P_{r \sim U(0,1)}(\mathbf{X}_{0}$,$\mathbf{X}_{L})$ & 99.27 & 80.66 & \textbf{96.00} & \textbf{69.11} \\ 
%  & $Q_{r \sim U(0,1)}(\mathbf{X}_{0}$,$\mathbf{X}_{L})$ & 99.27 & 81.26 & 95.23 & 68.82 \\ 
\hline
\multicolumn{2}{c|}{Ours ($\mathbf{X}_{0}$ and $\mathbf{X}_{L}$)} & 99.29 & \textbf{81.35} & 95.51 & 68.88 \\ 
\hline
\end{tabular}
}
\vspace{-5mm}
\end{center}
\label{table:select-images}
\end{table}

\section{Strategies for selecting images to augment}
\label{supp_sec3}

In the main paper, we provided a performance on a basic learning strategy using both original images $\mathbf{X}_{0}$ and augmented images $\mathbf{X}_{L}$. Table~\ref{table:select-images} shows various data fraction methods to effectively use augmented data for training. RandConv~\cite{xu2020robust} applied augmentation with half probability for every mini-batch. That is, sometimes the original images are used and at other times the augmented images are used for training. We call this a batch-level image augmentation strategy. We first compare these batch-level augmentation strategies using only the original images, using only the augmented images, and using both sets with half probability. Using only original images significantly degrades out-of-domain performance. On the other hand, using only augmented images degrades in-domain performance, especially in PACS. Therefore, it is important to properly combine the two types of images to balance in-domain and out-of-domain performance.

Next, we provide experiments on an instance-level augmentation strategy to learn both original and augmented images within a mini-batch. $P_{r}(\mathbf{X}_{0}, \mathbf{X}_{L})$ indicates this strategy, where $r$ is the data fraction of the original images. The larger $r$, the higher the proportion of the original images in the mini-batch. Generally, a high value of $r$ tends to improve in-domain performance and decrease out-of-domain performance. The most appropriate solution is to set $r$ to a value of 0.5 or to sample $r$ from $U(0,1)$. In particular, the random sampling strategy achieves satisfactory values for both in-domain performance and out-of-domain performance, and obtains comparable performance to the basic strategy using both original and augmented images. It is noteworthy that RandConv degrades the in-domain performance on PACS from 95.37\% to 92.75\% compared to the baseline, whereas our method improves both in-domain and out-of-domain performance.

\begin{table}[t]
\caption{Performance analysis for detailed components in terms of accuracy (\%). LeNet and ResNet18 are used for training on Digits and PACS, respectively. SDG and MDG indicate single domain generalization and multi domain generalization settings, respectively. \textit{Single} denotes the single-layer approach used by RandConv. \textit{Multi} (D/S) represents our progressive approach, where D means to initialize all layers differently, and S means to initialize one layer and use it equally for all layers.}
\begin{center}
\vspace{-5mm}
\scalebox{0.80}{
\begin{tabular}{c|ccc|c|cc}
\hline
\multirowcell{2}{Model}  & \multirowcell{2}{Conv. \\ smooth} & \multirowcell{2}{Contrast} & \multirowcell{2}{Offsets} & Digits & \multicolumn{2}{c}{PACS} \\
\cline{5-7} 
 &  &  &  & SDG & SDG & MDG \\
\hline
Baseline & - & - & -                                   & 52.00 & 63.13 & 81.45 \\
\hline
\multirowcell{6}{\textit{Single}} & - & - & -       & 74.84 & 67.50 & 82.43 \\
  & \checkmark & - & -       & 74.34 & 67.95 & 82.53 \\
  & - & \checkmark & -       & 77.14 & 67.81 & 83.16 \\
  & - & - & \checkmark       & 75.73 & 67.24 & 82.38 \\
  \cline{2-7} 
  & \multicolumn{3}{c|}{$w \sim N(0,\sigma_w), \sigma_w \sim U(\epsilon,1)$} & 76.59 & 67.46 & 82.53 \\
  & \multicolumn{3}{c|}{$w \sim N(0,\sigma_w), \sigma_w \sim U(\epsilon,2)$} & 75.80 & 67.99 & 82.50 \\
\hline
\textit{Multi} (D)      & - & - & -                                         & 74.72 & 66.47 & 82.49 \\
\hline
\multirowcell{8}{\textit{Multi} (S)} & - & - & -                            & 78.26 & 67.89 & 83.72 \\
  & - & \checkmark & -                                                      & 77.09 & 68.73 & 84.17 \\
  & - & - & \checkmark                                                      & 77.41 & 68.25 & 84.04 \\
  & - & \checkmark & \checkmark                                             & 77.08 & \textbf{69.01} & 84.24 \\
  \cline{2-7} 
  & \checkmark & - & -                                                      & 80.03 & 68.3 & 83.79 \\
  & \checkmark & \checkmark & -                                             & 80.02 & 68.55 & 84.22 \\
  & \checkmark & - & \checkmark                                             & 81.06 & 67.98 & 83.77 \\
  & \checkmark & \checkmark & \checkmark                                    & \textbf{81.35} & 68.88 & \textbf{84.29} \\
\hline
\end{tabular}
}
\vspace{-5mm}
\end{center}
\label{table:component}
\end{table}

\section{Component analysis}
\label{supp_sec4}

Table~\ref{table:component} shows a detailed performance comparison for each component of Pro-RandConv. First, we analyze whether we can improve performance by adding our components to the single-layer approach used in RandConv~\cite{xu2020robust}. Gaussian smoothing of convolution weights does not have a significant effect in a single-layer approach, whereas contrast diversification and deformable offsets help to improve performance. However, it does not contribute to a significant performance improvement, because of the limitation of style diversity and the problem of excessive semantic distortion in the single-layer approach. In addition, the method of variously adjusting the variance of the Gaussian distribution without fixing the convolution weight to He-initialization~\cite{he2015delving} shows some performance improvement on Digits.

Second, we analyze the influence of components in detail under our progressive approach. The key to the progressive approach is to initialize one layer and keep the remaining layers with the same parameters, which leads to a significant performance improvement. Next, we compare the performance with and without Gaussian smoothing of the convolution layer. In Digits, since the size of the object is relatively small, the multi-layer structure of the $3\times3$ convolution layer has excessive diversity. Thus, increasing the contrast and texture diversity without Gaussian smoothing has the effect of inducing semantic distortion. In other words, it is more effective to secure the contrast and texture diversity while controlling the deformation scale of texture with Gaussian smoothing. Conversely, in PACS, since the resolution of the image is large, the multi-layer structure of the $3\times3$ convolution layer is inefficient in diversity. Therefore, even if Gaussian smoothing is not applied, the generalization capability can be improved by contrast diversification and deformable offsets.

\section{Additional performance analysis}
\label{supp_sec5}

\subsection{Comparison with traditional augmentation}

In this section, we compare the traditional augmentation methods with our Pro-RandConv. Table~\ref{table:aug-digits} and Table~\ref{table:aug-pacs} provide performance comparisons on Digits and PACS, respectively. In both datasets, \textit{color jitter} and \textit{grayscale} are more effective than \textit{perspective} and \textit{rotate} in terms of improving generalization ability. Also, AutoAugment~\cite{Cubuk_2019_CVPR} and RandAugment~\cite{cubuk2020randaugment}, which apply various augmentation types simultaneously, enhance domain generalization capability more than single augmentation methods. Furthermore, the proposed Pro-RandConv outperforms all these augmentation methods with a simple random network structure. Thanks to its effective generalization capability, we argue that the proposed Pro-RandConv could be a strong baseline for various tasks.

\begin{table}[t]
\caption{Performance comparison with traditional augmentation techniques in the single domain generalization setting on Digits in terms of accuracy (\%). Each column title indicates the target domain. LeNet is used for training. * denotes reproduced results.}
\begin{center}
\vspace{-5mm}
\scalebox{0.80}{
\begin{tabular}{l|cccc|c}
\hline
Methods & SVHN & MNIST-M & SYN & USPS & Avg. \\
\hline
Baseline                                    & 32.52 & 54.92 & 42.34 & 78.21 & 52.00 \\
\hline
Color jitter*                               & 36.04 & 57.56 & 43.94 & 77.76 & 53.83 \\
Grayscale*                                  & 32.92 & 55.44 & 42.38 & 78.22 & 52.24 \\
Pespective*                                 & 33.63 & 43.86 & 40.92 & 69.12 & 46.88 \\
Rotate*                                     & 31.99 & 54.86 & 38.22 & 69.54 & 48.65 \\
\hline
AutoAugment~\cite{Cubuk_2019_CVPR}         & 45.23 & 60.53 & 64.52 & 80.62 & 62.72 \\
RandAugment~\cite{cubuk2020randaugment}    & 54.77 & 74.05 & 59.60 & 77.33 & 66.44 \\
\hline
\textbf{Ours}    & \textbf{69.67} & \textbf{82.30} & \textbf{79.77} & \textbf{93.67} & \textbf{81.35} \\
\hline
\end{tabular}
}
\vspace{-3mm}
\end{center}
\label{table:aug-digits}
\end{table}

\begin{table}[t]
\caption{Performance comparison with traditional augmentation techniques in the single domain generalization setting on PACS in terms of accuracy (\%). Each column title indicates the source domain. ResNet18 is used for training. * denotes reproduced results.}
\begin{center}
\vspace{-5mm}
\scalebox{0.80}{
\begin{tabular}{l|cccc|c}
\hline
Methods & Art & Cartoon & Photo & Sketch & Avg. \\
\hline
Baseline                                    & 74.64 & 73.36 & 56.31 & 48.27 & 63.15 \\
\hline
Color jitter*                               & 75.94 & 76.56 & 59.27 & 50.24 & 65.50 \\
Grayscale*                                  & 74.29 & 75.75 & 58.96 & 47.67 & 64.17 \\
Pespective*                                 & 72.29 & 70.17 & 59.99 & 43.79 & 61.31 \\
Rotate*                                     & 73.47 & 71.06 & 56.95 & 46.61 & 62.02 \\
\hline
AutoAugment*~\cite{Cubuk_2019_CVPR}         & 76.48 & 77.09 & 60.99 & 52.46 & 66.76 \\
RandAugment*~\cite{cubuk2020randaugment}    & 76.76 & 78.00 & 62.09 & 56.40 & 68.31 \\
\hline
\textbf{Ours}    & \textbf{76.98} & \textbf{78.54} & \textbf{62.89} & \textbf{57.11} & \textbf{68.88} \\
\hline
\end{tabular}
}
\vspace{-5mm}
\end{center}
\label{table:aug-pacs}
\end{table}

\begin{table}[t]
\caption{Performance comparison on Digits in detail for a fair comparison (\%). In MNIST-M, two different kinds of sets (A/B) are utilized. LeNet is used for training.  Ours${}^{-T}$ and Ours${}^{-C}$ indicate disabling texture diversification and contrast diversification, respectively. RC denotes the official results of RandConv.}
\begin{center}
\vspace{-5mm}
\scalebox{0.75}{
\begin{tabular}{l|cccc|c}
\hline
Methods & SVHN & MNIST-M (A/B) & SYN & USPS & Average (A/B) \\
\hline
RC~\cite{xu2020robust} & 57.52 & \quad - \quad / 87.76 & 62.88 & 83.36 & \quad - \quad / 72.88 \\
\hline
Ours${}^{-T}$               & 62.76 & 74.52 / 81.91 & 78.07 & 93.01 & 77.09 / 78.94 \\
Ours${}^{-C}$               & \textbf{70.35} & \textbf{82.98 / 88.34} & 77.40 & 93.52 & 81.06 / 82.40 \\
\textbf{Ours}               & 69.67 & 82.30 / 87.72 & \textbf{79.77} & \textbf{93.67} & \textbf{81.35} / \textbf{82.72} \\
\hline
\end{tabular}
}
\vspace{-3mm}
\end{center}
\label{table:mnist-m}
\end{table}

\subsection{Fair comparison on MNIST-M}

We confirmed that RandConv uses the test set of MNIST-M~\cite{DANN} differently from the existing methods (\eg PDEN~\cite{li2021progressive}, M-ADA~\cite{qiao2020learning}, and ME-ADA~\cite{zhaoNIPS20maximum}). Existing methods use MNIST-M consisting of 9,001 images, which we refer to as set A. RandConv uses MNIST-M which consists of 10,000 images, which we refer to as set B. For a fair comparison, we compare the performance of both MNIST-M sets. Table~\ref{table:mnist-m} shows that performance comparison on two sets of MNIST-M. We emphasize that our Pro-RandConv method has higher generalization capability than RandConv~\cite{xu2020robust} in all domains including MNIST-M.

\begin{figure*}[t!]
\centering
\includegraphics[width=\linewidth]{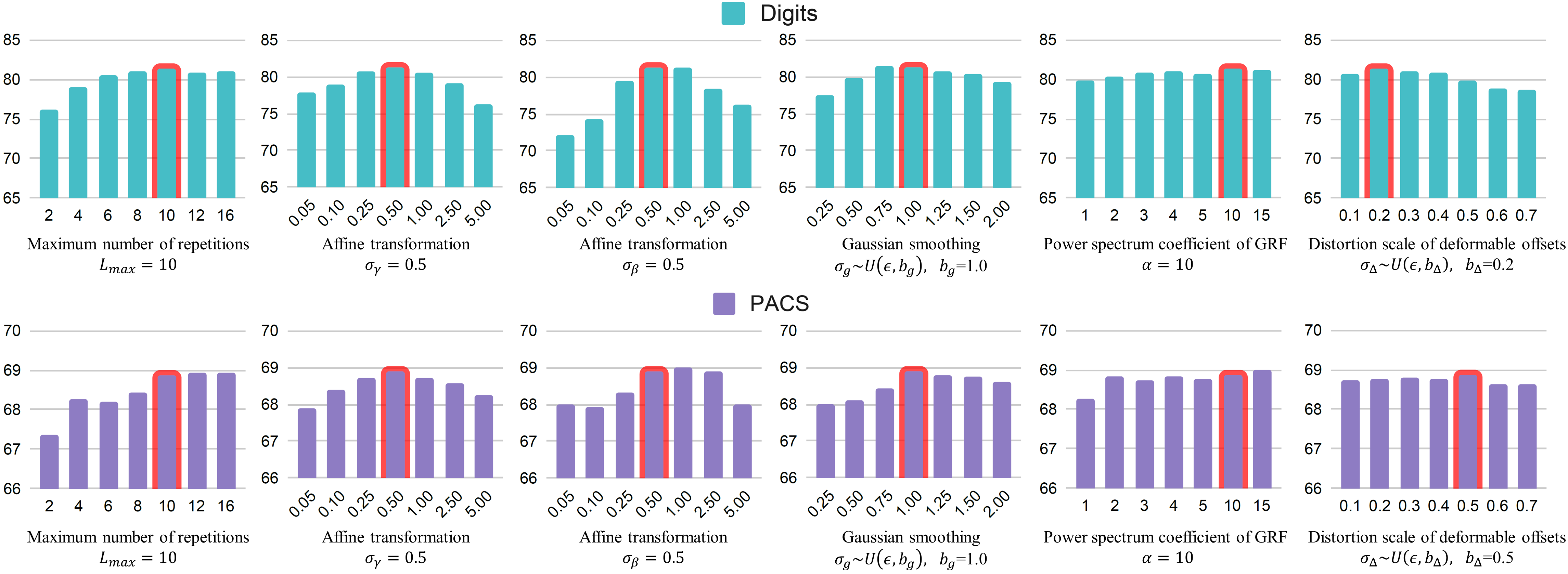}
\caption{Analysis of hyperparameter selection in the single domain generalization setting on Digits and PACS.}
\label{fig:hparam}
\end{figure*}

\section{Hyperparameter selection}
\label{supp_sec6}

\subsection{Hyperparameters of the progressive approach}

The core idea of this paper is a progressive method that initializes a random convolution layer once and then stacks it multiple times with the same structure. Eventually, from a hyperparameter selection perspective, RandConv's traditional approach of choosing the kernel size changes to choosing the number of repetitions of the convolution layers. For example, RandConv generates random-style images based on a kernel size randomly selected from $\{1,3,5,7\}$ for each mini-batch. In a similar way, we choose a different number of repetitions with uniform sampling from 1 to $L_{max}$ for each mini-batch. Figure 1(c) and 2(a) in the main paper show that as the kernel size increases, images augmented by RandConv easily lose their semantics and eventually the performance degrades rapidly. The progressive approach, on the other hand, is less sensitive to increasing $L_{max}$, since the performance does not degrade significantly as the receptive field increases, as shown in Fig~\ref{fig:hparam}. However, the computational cost increases proportionally to the number of repetitions, so we chose a reasonable value of 10 to account for the tradeoff. 

\subsection{Hyperparameters of convolution blocks}

We further provide a performance comparison for all hyperparameters in the random convolution block, as shown in Fig.~\ref{fig:hparam}. We first analyze the hyperparameters for contrast diversification. We chose $\sigma_\gamma$ and $\sigma_\beta$ to be 0.5, as they show the highest performance on both Digits and PACS datasets. This means that the affine transformation parameters, $\gamma$ and $\beta$, are sampled from $N(0, 0.5^2)$. Figure 7(a) and (b) in the main paper show that $\gamma$ and $\beta$ can cause false distortion or saturation if they are smaller or larger than 0.5, so we recommend keeping them at 0.5 regardless of the dataset. 

Next, we analyze the hyperparameters for the convolution weights. The convolution weights are initialized by~\cite{he2015delving} as in RandConv (i.e., $\sigma_{w}=1/\sqrt{k^2 C_{in}}=1/\sqrt{3^3}$). We further apply Gaussian smoothing to this kernel. For Gaussian smoothing $g[i_m,j_m] = \exp(-\frac{i_m^2+j_m^2}{2\sigma^2_g})$, the smoothing scale is sampled from $\sigma_g \sim U(\epsilon, b_g)$, where $\epsilon$ indicates a small value. This means that $\sigma_g$ is randomly sampled for each mini-batch, so the smoothing effect is different each time. This technique can be used to mitigate the problem of severely distorted object semantics when the random offset of the deformation convolution is too irregular and large in scale. We chose $b_g$ to be 1.0 because it performs best on both Digits and PACS datasets. As with the hyperparameter selection for contrast diversification, we set the same value for all datasets. 

Finally, we introduce hyperparameters for deformable convolution that further enhance texture diversity. The tensor for deformable offsets consists of $(2k^2, H, W)$, where $k$ is the kernel size of the convolution layer, and $H$ and $W$ are the height and weight of an image, respectively. That is, there are a total of $2k^2$ offsets per pixel in the image of $H \times W$, where 2 means the values of $\Delta i_m$ and $\Delta j_m$. To induce natural geometric variation, we consider spatial correlation by generating a total of $2k^2$ Gaussian Random Fields (GRF) with a size of $H \times W$. We refer to this code\footnote{https://github.com/bsciolla/gaussian-random-fields} for the GRF implementation, where spatial correlation can be controlled by varying the coefficient $\alpha$ of the power spectrum. As shown in Fig. 7(f) of the main paper, the larger the coefficient $\alpha$, the higher the spatial correlation. We scaled the Gaussian random field (GRF) by choosing a coefficient of 10 for the power spectrum. Another hyperparameter is the distortion scale $\sigma_\Delta$ of the deformable offset. In particular, geometric information such as rotation is an important attribute for digits recognition, so severe deformation impairs class-specific semantic information. Figure 7(e) in the main paper shows that the shape of the object becomes unrecognizable as the scale increases. This hyperparameter is also related to the size of the image, so we choose different hyperparameters according to image size. For Digits, a small scale of 0.2 is used, while for PACS, OfficeHome, and VLCS, a scale of 0.5 is used. As with the other hyperparameters, uniform sampling is performed as $U(\epsilon, b_{\Delta})$ to make it less sensitive to hyperparameter selection.

\end{document}